\documentclass{article} 
\usepackage{iclr2022_conference,times}

\usepackage{amsmath,amsfonts,bm}

\def\eqref#1{equation~\ref{#1}}

\def\1{\bm{1}}

\DeclareMathAlphabet{\mathsfit}{\encodingdefault}{\sfdefault}{m}{sl}
\SetMathAlphabet{\mathsfit}{bold}{\encodingdefault}{\sfdefault}{bx}{n}

\usepackage{multirow}
\usepackage{hyperref}
\usepackage{url}
\usepackage{caption}
\usepackage{subcaption}
\usepackage{graphicx}
\usepackage{booktabs} 
\usepackage{tabularx}
\usepackage{enumitem}
\usepackage{wrapfig}
\usepackage{titlesec}
\usepackage{float}

\newcommand\blfootnote[1]{%
  \begingroup
  \renewcommand\thefootnote{}\footnote{#1}%
  \addtocounter{footnote}{-1}%
  \endgroup
}
  
\title{Sequence-to-Sequence Modeling for Action Identification at High Temporal Resolution}

\author{Aakash Kaku$^{*1}$, Kangning Liu$^{*1}$, Avinash Parnandi$^{*2}$, Haresh Rengaraj Rajamohan$^{1}$, \\ 
  \textbf{Kannan Venkataramanan$^{1}$,  Anita Venkatesan$^{2}$,  Audre Wirtanen$^{2}$, Natasha Pandit$^{2}$,} \\
  \textbf{Heidi Schambra$^{\dagger 2}$, Carlos Fernandez-Granda$^{\dagger 1,3}$} \\
  $^1$NYU Center for Data Science $^2$NYU School of Medicine \\
  $^3$ Courant Institute of Mathematical Sciences
  }

\iclrfinalcopy 
\begin{document}

\maketitle

\begin{abstract}
Automatic action identification from video and kinematic data is an important machine learning problem with applications ranging from robotics to smart health. Most existing works focus on identifying coarse actions such as running, climbing,  or cutting a vegetable, which have relatively long durations. This is an important limitation for applications that require identification of subtle motions at high temporal resolution. For example, in stroke recovery, quantifying rehabilitation dose requires differentiating motions with sub-second durations. Our goal is to bridge this gap. To this end, we introduce a large-scale, multimodal dataset, StrokeRehab, as a new action-recognition benchmark that includes subtle short-duration actions labeled at a high temporal resolution. These short-duration actions are called functional primitives, and consist of reaches, transports, repositions, stabilizations, and idles. The dataset consists of high-quality Inertial Measurement Unit sensors and video data of 41 stroke-impaired patients performing activities of daily living like feeding, brushing teeth, etc. We show that current state-of-the-art models based on segmentation produce noisy predictions when applied to these data, which often leads to overcounting of actions. To address this, we propose a novel approach for high-resolution action identification, inspired by speech-recognition techniques, which is based on a sequence-to-sequence model that directly predicts the sequence of actions. This approach outperforms current state-of-the-art methods on the StrokeRehab dataset, as well as on the standard benchmark datasets 50Salads, Breakfast, and Jigsaws.
\end{abstract}
\blfootnote{$^\ast$Equal contribution, $^\dagger$Joint last authors. Correspondence to Aakash Kaku (\texttt{ark576@nyu.edu}) or Kangning Liu (\texttt{kl3141@nyu.edu}).}

\section{Introduction}
\label{sec:introduction}
Automatic action identification from video and kinematic data is an important machine learning problem with applications ranging from robotics to smart health. Most previous works (e.g. TCN~\citep{lea2017temporal}, MS-TCN~\citep{farha2019ms}, or ASRF~\citep{ishikawa2021alleviating}) focus on identifying coarse actions such as running, climbing, sitting, putting bread on a plate, cutting a tomato, drinking from a glass, or inserting a needle, which usually have long durations. 
An important question is whether these approaches are effective in applications that require identification of short, single-goal motions at high temporal resolution, such as the rehabilitation of stroke patients. Identifying subtle, short-duration actions is key in data-driven stroke rehabilitation. 
Basic research in animals indicates that the repeated practice of functional motions early after stroke markedly boosts recovery~\citep{jeffers2018does,bell2015training,murata2008effects}. The same is believed to be true for humans undergoing rehabilitation for stroke-induced disability. However, there has been no systematic quantification of how many repetitions are needed early after stroke for optimal recovery~\citep{krakauer2012getting}. Data-driven quantification requires identifying and counting motions at high temporal (sub-second) resolution.



In order to advance methodology for high-resolution action identification, it is crucial to establish appropriate benchmark datasets. Existing benchmarks, such as 50Salads~\citep{stein2013combining}, Breakfast~\citep{kuehne2014language}, Jigsaws~\citep{gao2014jhu}, Kinetics~\citep{kay2017kinetics} or FineGym~\citep{shao2020finegym}, contain very few short-duration actions (see Figure~\ref{fig:dist_action_duration}). To address this,
we introduce a large-scale, multimodal dataset, StrokeRehab, as a new action-recognition benchmark that includes subtle short-duration actions labeled at a high temporal resolution. The dataset consists of high-quality wearable sensor and video data of 41 patients who are mildly or moderately impaired by stroke. These patients performed nine activities of daily living like drinking, eating, applying deodorant, etc. in a rehabilitation gym. The subtle actions performed by the patients in each session were meticulously labeled by trained annotators overseen by an expert, who examined one-third of their labels. 

Evaluation of current state-of-the-art approaches for action recognition~\citep{farha2019ms,ishikawa2021alleviating}) on the StrokeRehab dataset reveals a limitation of these methods for high-resolution action identification. 
These approaches are based on segmentation; they assign an action to each individual time step of the input data. It has been previously reported that this often yields noisy estimates~\citep{ishikawa2021alleviating}. Current methods address this issue by promoting smoothness of the segmentation output~\citep{farha2019ms,li2020ms} or by separately identifying the boundaries between actions~\citep{ishikawa2021alleviating}.  
We observe that these approaches are not as effective when applied to the short-duration subtle actions in the StrokeRehab dataset, where boundaries are not clearly defined even for human-expert annotators. As a result, segmentation-based approaches produce noisy estimates, which limits their accuracy. To address this limitation, we introduce a novel approach to action identification inspired by speech-recognition models.


We propose a sequence-to-sequence (seq2seq) model that estimates the sequence of actions directly, without attempting to identify the boundaries between actions. The model learns to map variable-length input to variable-length output sequences using an encoder and a decoder. The encoder generates a hidden state vector from the input data, which is then decoded to generate an estimated sequence of actions. Such sequence-to-sequence approaches have been highly effective in speech recognition and natural language applications~\citep{chan2015listen,chiu2018state,prabhavalkar2017comparison}. The sequence-to-sequence model achieves state-of-the-art performance on the StrokeRehab dataset, and also outperforms existing approaches on the standard benchmark datasets like Breakfast, 50Salads, and Jigsaws for the task of sequence-prediction. 

In summary, the contributions of our work are the following:
\begin{itemize}[leftmargin=*]
    \item We release a large-scale, multimodal dataset - StrokeRehab - which can be used for training and evaluating models to identify subtle actions at a high temporal resolution.
    \item We show that existing state-of-the-art methods based on segmentation generate noisy predictions when applied to the subtle short-duration actions in StrokeRehab, which leads to overcounting. 
    \item Taking inspiration from speech recognition models, we propose a sequence-to-sequence method to predict a sequence of actions rather than individual, segmented actions. The proposed model outperforms the existing state-of-the-art methods on the StrokeRehab dataset and the benchmark datasets: 50Salads, Breakfast, and Jigsaws.
\end{itemize}

\section{Stroke rehabilitation dataset}
\label{sec:stroke_rehab_dataset}
\subsection{Clinical motivation}
Stroke is the leading cause of disability in the United States. It affects nearly one million individuals per year, with the numbers of stroke cases increasing as our population ages \citep{intro1,intro2,broderick2004william}. Stroke affects the arm in 77\% of patients, causing long-lasting motor impairment~\citep{lawrence2001estimates,kwakkel2015constraint}. By six months, most of these patients remain unable to independently perform activities of daily living, such as feeding, bathing, grooming, etc. This loss of independence reduces the quality of life of both the patients and their caretakers~\citep{nichols2005factors,carod2009quality} and exacts a heavy societal toll, with caretaking and healthcare costs predicted to skyrocket to \$240 billion by 2030~\citep{heidenreich2011forecasting}. Due to the profound impact of the stroke on the arm and its downstream consequences, we focus on the arm in our study. 

Following stroke, some spontaneous recovery occurs because of brain plasticity, but this plasticity alone does not fully restore function. In animal models of stroke, training high numbers of functional arm motions not only increases this plasticity~\citep{kim2018coordinated,bell2015training}, but also markedly boosts recovery~\citep{murata2008effects,jeffers2018does}. It is increasingly believed that if intensive rehabilitation training can be delivered early after stroke in humans, recovery could be similarly accelerated~\citep{krakauer2012getting}.

\begin{figure}[h]
\centering
\includegraphics[width=0.9\textwidth]{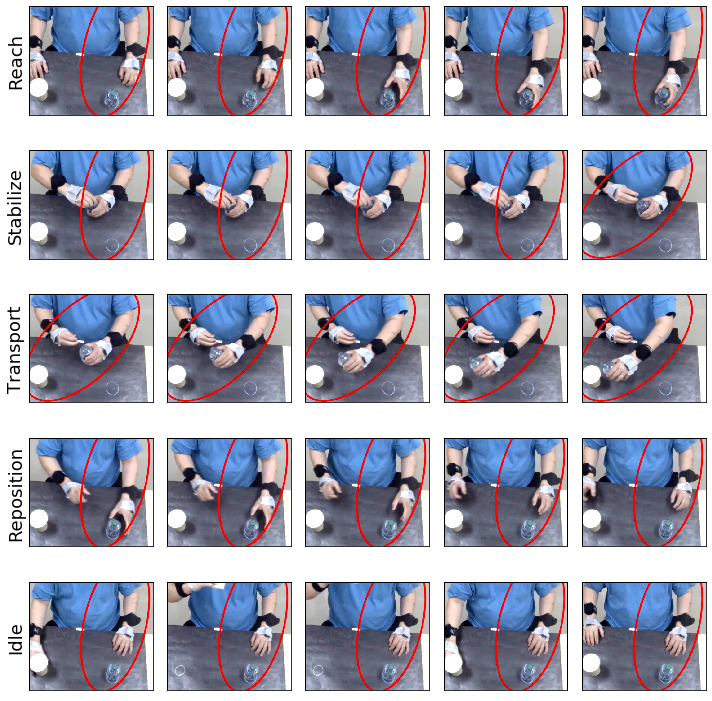}
\caption{A stroke patient performing a functional activity (drinking) from the  StrokeRehab activities battery. Using the functional motion taxonomy, the activity can be decomposed into its constituent functional primitives as follows: \emph{reach}, upper extremity (UE) motion to bring it into contact with a target object (e.g. water bottle); \emph{stabilize} minimal UE motion to keep a target object still (e.g. holding the bottle to allow the other UE to open the cap); \emph{transport}, UE motion to move a target object (e.g. moving the bottle to pour some water); \emph{reposition}, UE motion proximate to a target object (e.g. to move the to the initial neutral spot); \emph{idle}, minimal UE motion to stand at the ready near a target object.}
\label{fig:primitives_examples}
\end{figure}

In rehabilitation training, patients use their impaired arm to practice activities of daily living (ADLs). ADLs are composed of five fundamental motions called functional primitives: reach, transport, reposition, stabilize, and idle~\citep{schambra2019taxonomy}. For example, in a drinking activity, our arm would "idle" as it rests at our side, then "reach" for a glass and "transport" the glass to our mouth, then "transport" the glass back to the table, and finally "reposition" back to our side for an "idle." A visual example of primitives can be seen in Figure~\ref{fig:primitives_examples}.

A major clinical question is how many repetitions of functional motions are needed for optimal recovery. In animal research, the number of motions that boost arm recovery has been  quantified~\citep{jeffers2018does}. For humans, this quantification has not been done. A handful of studies have observed that patients train about 10 times fewer repetitions than what recovering animals receive, suggesting pronounced under-training in human rehabilitation ~\citep{lang2009observation,kimberley2010comparison}. However, the optimal number of training repetitions to boost recovery remains uncertain in humans. Currently, the best way to quantify rehabilitation is hand tallying. If performed in real time by the rehabilitation therapist, hand tallying distracts from treatment delivery. If the session is videotaped and annotated offline, tallying is laborious and slow: \textbf{it takes one hour of manual effort to label one minute of recorded training}. Hand tallying thus incurs time, effort, and personnel costs that render it unscalable. 

Therefore, to facilitate the quantification of functional motions performed during stroke rehabilitation, we developed an approach that combines unobtrusive motion capture with automated identification. We collected the StrokeRehab dataset that consists of labeled sensor and video data from stroke patients. We then used the StrokeRehab dataset to train models to automatically identify and count functional primitives. We will make the complete dataset publicly available online upon publication (see here for sample data\footnote{\url{https://drive.google.com/drive/folders/1_a48XeRjFRdwaiaQXAV3tvAYb-4VmDbM?usp=sharing}}).  

\subsection{Cohort selection} 
We collected sensor and video data from 41 stroke patients in an inpatient rehabilitation gym. Individuals were included if they were $\geq$ 18 years old, had premorbid right-handed dominance, and had unilateral arm weakness from ischemic or hemorrhagic stroke. Details about the inclusion and exclusion criteria and patient demography are provided in Appendix~\ref{app:sub_sec_cohort_selection}. 

\subsection{Data Acquisition and Labelling}
\label{sec:data}
Upper body motion was recorded while the patients performed activities of daily living commonly used during stroke rehabilitation. The activities included: washing the face, applying deodorant, combing the hair, donning and doffing glasses, preparing and eating a slice of bread, pouring and drinking a cup of water, brushing teeth, and moving an object horizontal and vertical target arrays. See Appendix~\ref{sec:app_desc_act} for detailed descriptions of the activities. The patients performed five repetitions of each activity.

\textbf{Description of kinematic data}: Upper body motion was recorded using nine Inertial Measurement Units (IMUs, Noraxon, USA) attached to the upper body, specifically the cervical vertebra C7, the thoracic vertebra T12, the pelvis, and both arms, forearms, and hands. These IMUs captured 76-dimensional kinematic features of 3D linear accelerations, 3D quaternions, and joint angles from the upper body (see Appendix~\ref{app_subsec:kinematic_data} and Appendix~\ref{sec:app_joint_angles} for details). As an additional feature, we included the paretic (stroke-affected) side of the patient (left or right) encoded in a one-hot vector, increasing the dimension of the feature vector to 77.

\textbf{Description of video data}: Video data were synchronously captured using two high definition cameras (1088 x 704, 60 frames per second; Ninox, Noraxon) placed orthogonally $<$ 2 m from the patient. We also extracted frame-wise feature vectors from the raw videos. The detailed procedure to extract these feature vectors is mentioned in Appendix~\ref{app_subsec:video_feat_ext}.

\textbf{Data labeling}: The subtle actions performed by the patients in each session were meticulously labeled by trained annotators overseen by an expert, who examined one-third of their labels. Interrater reliability between the coders and expert was high, with Cohen’s kappa $\geq$ 0.96 between the coders and the expert. Details on the labeling procedure are provided in Appendix~\ref{app_subsec:data_labeling}

\textbf{Train and test set}:
We assigned 33 patients to a training set and 8 patients to a test set. To this end, the 41 patients were separated into eight subgroups, balancing for impairment level and paretic side (left or right). One patient in each group was randomly removed and assigned to the test set. The remaining patients were assigned to the training set.

\subsection{Data statistics and comparison to benchmark action recognition datasets}
The dataset consists of 1,763 trials of rehabilitation activities performed by 41 stroke-impaired
patients. Cumulatively, they performed 64,284 functional primitives, making this dataset one of the largest in terms of the number of annotated actions. In comparison, the previous benchmark datasets like Breakfast (11,656 annotated actions), Jigsaws (1,701 annotated actions), and 50Salads (999 annotated actions) have a relatively low number of labeled actions. A non-trivial amount of manual effort (approximately 4,320 human hours) was required to label the 64,284 functional primitives.

\begin{figure}
\begin{minipage}{0.49\textwidth}
\centering
\includegraphics[width=\textwidth]{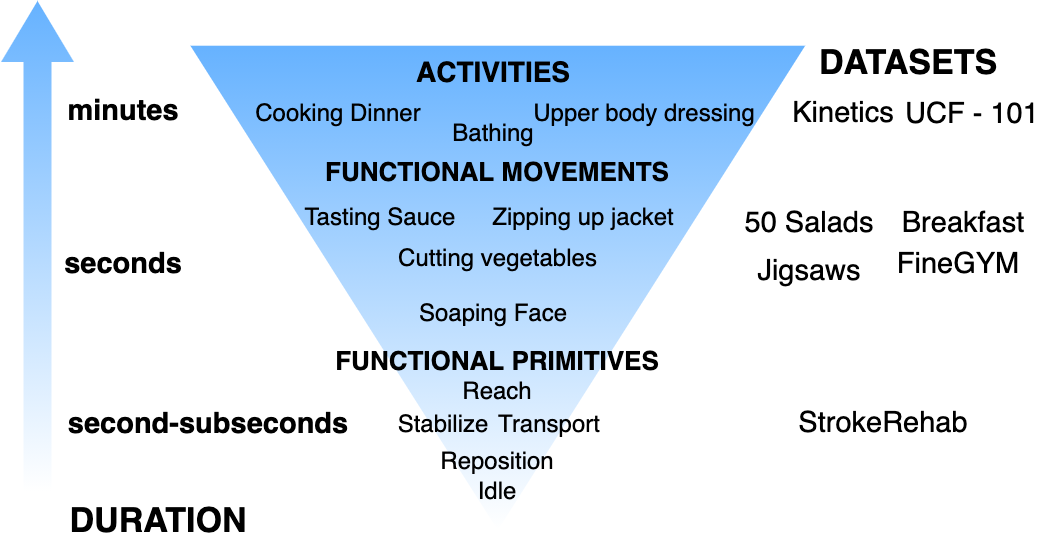}
\caption{Popular action recognition datasets are aligned to a hierarchy of the labeled actions they contain. Our dataset StrokeRehab mainly contains short-duration basic functional primitives.}

\label{fig:tax_action}
\end{minipage}\hfill
\begin{minipage}{0.49\textwidth}
\centering
\includegraphics[width = 0.9\textwidth]{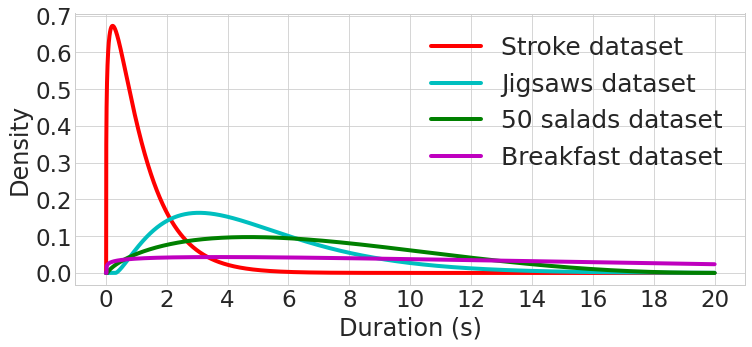}
\caption{Distribution of action duration of actions for the various benchmark datasets. This illustrates the extreme fine-grained nature of actions (functional primitives) in the StrokeRehab dataset in comparison to existing ones.}
\label{fig:dist_action_duration}
\end{minipage}
\end{figure}

In Figure~\ref{fig:tax_action}, we show that the actions in StrokeRehab are at the bottom of the action hierarchy~\citep{schambra2019taxonomy}. They are functional primitives with short durations that execute one goal. In existing datasets, most actions are complex functional movements or activities, which have long durations and execute several goals. The difference in duration is evident in Figure~\ref{fig:dist_action_duration}. For the StrokeRehab dataset, most actions have a time duration of less than a second, whereas in the other datasets most actions take several seconds.

\section{Relevant work}
\label{sec:relevant_work}

\textbf{Action classification}:
Action classification is the task of identifying a single action, as opposed to a sequence of actions. 
Several methods 
use 2D CNNs to extract frame-wise features from an input video, which are then combined to predict a coarse action taking place in the video~\citep{wang2018temporal,lin2019tsm,zhou2018temporal}.
Alternatively, techniques such as  C3D~\citep{tran2015learning},  I3D~\citep{carreira2017quo} SlowFast~\citep{feichtenhofer2019slowfast} and X3D~\citep{feichtenhofer2020x3d} use 3D CNNs to exploit the spatial-temporal information in the data. 
There also exist several works that perform action classification from kinematic data
~\citep{attal2015physical,chung2019sensor}. 

\textbf{Action segmentation}: Action segmentation is the problem of segmenting an input stream of data, labeling each frame according to the action that is being carried out.
Earlier methods for action segmentation employed hidden Markov models~\citep{kuehne2016end,guerra2017capture}. More recently, convolutional neural networks~\citep{wen2019time,kaku2020towards} and recurrent neural networks~\citep{singh2016multi} have been applied to this problem
Inspired by the success of temporal convolutional networks (TCNs) in speech synthesis, \cite{lea2017temporal} adapted these models to action segmentation.  
MS-TCN~\citep{farha2019ms}, which uses a multi-stage TCN architecture, has become one of the most widely used architecture for action segmentation. Although these methods achieve high frame-wise accuracy,  they still produce a significant number of over-segmentation errors. In order to address this, several boundary-aware methods have been developed which perform temporal smoothing of the frame-wise predictions~\citep{wang2020boundary,ishikawa2021alleviating}. These methods use ground-truth boundary information to train a binary classification network to perform boundary detection. The boundary estimates are then used to aggregate the frame-wise predictions either in a soft manner (boundary-aware pooling) or by setting a hard threshold. However, for subtle actions with a high temporal resolution, such as the functional primitives in the StrokeRheab dataset, the duration of each action is very short. As a result, the boundaries between actions can be hard to detect or even hard to define (see Figure~\ref{fig:boundary_time}). 

\textbf{Sequence-to-sequence models}:
Our proposed method is based on
sequence-to-sequence (Seq2seq) models.
These models allow us to learn a mapping of a variable-length input sequence to a variable-length output sequence~\citep{sutskever2014sequence}.
Popular sequence-to-sequence models include the RNN-transducer~\citep{graves2012sequence}, connectionist temporal classification (CTC) based models~\citep{graves2006connectionist} and encoder-decoder based models~\citep{chan2015listen}. Here we leverage encoder-decoder based models, which have very successful in machine translation~\citep{bahdanau2014neural,luong2014addressing}, speech recognition~\citep{chan2015listen} and image captioning~\citep{vinyals2015show}.

\section{Methods} \label{sec:methods}
\vspace{-3mm}

We consider the problem of estimating a sequence of actions from a sequence of input data. Let $\mathbf{x} = (x_1,...,x_T)$ be an input sequence with length $T$, which correspond to high-dimensional sensor or video features in our datasets of interest. We denote the corresponding sequence of actions by $\mathbf{y} = (y_1,...,y_{T'})$, where $y_i$ ($1\leq i \leq T'$) is one of $c$ different actions, so $y_i 
\in \{1,\ldots,c\}$. The length $T'$ of this sequence is much shorter than the input sequence because each action takes place over several time steps. 
For example, a 100-frame video sequence from the StrokeRehab dataset may correspond to the 3-action sequence: \emph{reach}, \emph{transport}, \emph{stabilize}. 

\begin{figure}[t]

\centering
\includegraphics[width=0.9\linewidth]{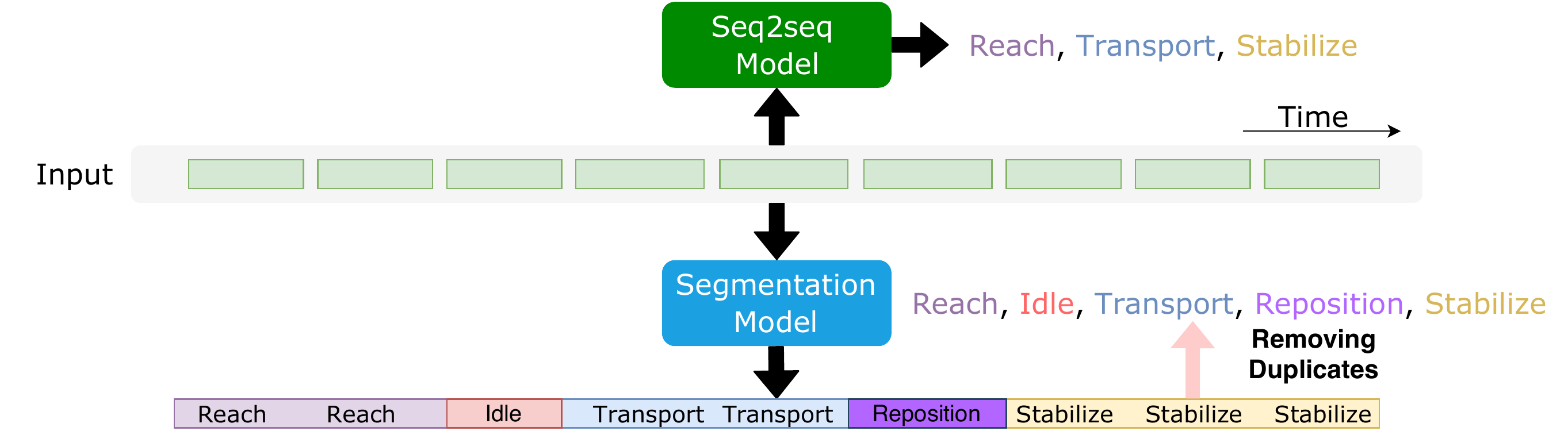}
\caption{Comparison of sequence-to-sequence (seq2seq) and segmentation models. The segmentation model outputs frame-wise action predictions, which can then be converted to a sequence estimate by removing the duplicates. The seq2seq model produces a sequence estimate directly.}
\label{fig:concept}
\end{figure}

\subsection{Action sequence prediction by action segmentation model}
\label{subsec:asm}

Most existing methods address the task of action sequence prediction by performing segmentation of the input data. Given $\mathbf{x}$, a segmentation model outputs frame-wise action labels $\mathbf{y_f} $ with the same length as the input. $\mathbf{y_f}$ can then be converted to $\mathbf{y}$ by removing label repetitions at consecutive steps (see Figure~\ref{fig:concept} for a concrete example). 
Segmentation-based models have been observed to systematically over-segment the input data
~\citep{farha2019ms,wang2020boundary,ishikawa2021alleviating}. \cite{wang2020boundary} and \cite{ishikawa2021alleviating} address this by training a separate action-boundary detection network. 
The boundary predictions are then used to refine the frame-wise predictions. As pointed out by~\cite{shao2020finegym}, boundaries of high-level actions are more detectable because the adjacent motions are distinctive; for example, the motions associated with cutting tomatoes versus tossing a salad are very different.
In contrast, boundaries of fine-grained actions are harder to identify because their transitions are more subtle; for example, the boundary between the end of a \emph{reach} and the beginning of a \emph{transport} in the StrokeRehab dataset is determined by when the finger pads have fully contacted a target object. 

Figure~\ref{fig:boundary_time} shows the accuracy of boundary-detection models for actions with different durations for several datasets. For all datasets, the accuracy of boundary detection is inversely proportional to duration. In addition, we observe that accurate boundary prediction can indeed be achieved for datasets with high-level actions (e.g. 50Salads and Breakfast), but is very challenging for the proposed dataset StrokeRehab. This suggests that segmentation-based approaches may be fundamentally limited for identification of subtle action sequences at high time resolution.

\subsection{Sequence-to-sequence modeling for action identification}
\label{subsec:seq2seq}

Motivated by the limitations of segmentation-based approaches described in the previous section, we propose to directly predict the sequence of actions from the input data using  sequence-to-sequence (seq2seq) models inspired by speech recognition (see Figure~\ref{fig:concept}). These models are designed to map input sequences to output sequences of different length, and are therefore well suited to the action-sequence estimation problem. The key idea is to \emph{encode} the input data as a hidden vector of fixed dimension. The hidden vector is then \emph{decoded} sequentially to produce the estimated sequence. The rest of this section explains the approach in detail.

\textbf{Encoding}: The encoder $f_{\text{enc}}$, which can be a convolutional or recurrent network, maps the input sequence $\mathbf{x}$ to a fixed-length hidden vector $\mathbf{h} (\theta_{\text{enc}}) = f_{\text{enc}}(\mathbf{x};\theta_{\text{enc}})$ ($\theta_{\text{enc}}$ are the parameters of the encoder). The hidden vector must capture any long-term dependencies in the input sequence, which can be challenging in the case of high-level actions with long durations. For such cases, we have observed that using features from a pre-trained segmentation model can boost performance (see Section~\ref{subsec:seq2seq imple}).

\textbf{Decoding}: The decoder is a recurrent neural network (RNN), which outputs the estimated sequence based on the hidden vector $\mathbf{h}$. For $i=1,2,\ldots$, the decoder estimates the conditional probability of the next action given the previous action and $\mathbf{h}$. To this end, the RNN $f$ maintains a decoder state $\mathbf{s_i}$, which is updated based on the previous action and the hidden vector:
\begin{align} \label{eqn:s_i}
 \mathbf{s_i}(\theta_{\text{dec}}) & := f_{\text{dec}}(\mathbf{s_{i-1}},\mathbf{h},y_{i-1};\theta_{\text{dec}}), 
 \end{align}
 where $\theta_{\text{dec}}$ denotes the parameters of the decoder. For $i=1$, the previous action is set to a \emph{start-of-sequence} token. The conditional probability of $y_i$ given the previous actions is then approximated using a multilayer perceptron (MLP) with a softmax output. Specifically,
\begin{align} \label{eqn:p_i}
\mathbf{p}_i(\theta_{\text{dec}}, \theta_{\text{enc}},\theta_{\text{mlp}}) := \text{Softmax}(\text{MLP}(\mathbf{s_{i}}, \mathbf{h};\theta_{\text{mlp}}))
\end{align}
is a $c+1$-dimensional vector that contains the estimates of the conditional probability that $y_i$ equals each of the $c$ possible actions or an \emph{end-of-sequence} token. 
The $i$th predicted action $\hat{y}_i$ is obtained by maximizing this conditional probability. The procedure continues until $\hat{y}_i$ equals the \emph{end-of-sequence} token. Optionally, we can incorporate an attention mechanism during decoding~\citep{bahdanau2014neural,chorowski2015attention}.

\textbf{Training}: For the sake of simplicity, we consider a single training example ($\mathbf{x},\mathbf{y}$), where the last entry of $\mathbf{y}$ contains the \emph{end-of-sequence} token.  
The parameters of the encoder and decoder networks are learned by maximizing the objective function:
\begin{equation}
   \max_{\theta_{\text{enc}},\theta_{\text{dec}},\theta_{\text{mlp}}} \sum_i \log(\mathbf{p}_i  [y_i]),
\end{equation}
where we omit the dependence of $\mathbf{p}_i$ on $\theta_{\text{enc}}, \theta_{\text{dec}}, \theta_{\text{mlp}}$ to ease notation. Here $\mathbf{p}_i  [y_i]$ denotes the probability that the model assigns to the true observed action $y_i$ when receiving $\mathbf{x}$ as an input.

During inference the ground-truth, $\mathbf{y}$ is not available, so the model can only use the previous predicted action  $\hat{y}_{i-1}$ to compute $\mathbf{s_i}$ in \eqref{eqn:s_i}. This suggests using the estimate also during training, a technique known as curriculum learning~\citep{bengio2015scheduled}. We apply this technique by replacing $y_{i-1}$ with $\hat{y}_{i-1}$ in~\eqref{eqn:s_i} with a probability $\epsilon$, which is increased gradually during training.

\textbf{Inference}: During inference, we perform greedy-decoding to find the most likely sequence of actions given the input data. Specifically, at each time step $i$, we use the previous prediction, $\hat{y}_{i-1}$, in \eqref{eqn:s_i} to compute the decoder state, $\mathbf{s_i}$, which in turn is used to compute $\mathbf{p_i}$ using \eqref{eqn:p_i}. 
Then, we choose $\arg \max_{1\leq j \leq c+1}\mathbf{p_i}[j]$ as the predicted action for step $i$ (note that there are $c+1$ possible actions because one is the \emph{end-of-sequence} token). Beam-search decoding is often preferred over greedy-decoding in speech recognition and natural language applications~\citep{chan2015listen}, but here it did not provide a significant improvement.

\section{Experiments}
\label{sec:experiments} 
\vspace{-3mm}
In our experiments, we systematically compared the performance of our proposed sequence-to-sequence model to existing segmentation-based approaches on the StrokeRehab dataset described in Section~\ref{sec:stroke_rehab_dataset}, as well as the  benchmark datasets 50Salads, Breakfast and Jigsaws (described in Appendix Appendix~\ref{app_subsec:datasets}). 

\subsection{Performance metrics}
\label{sec:metrics}
In order to evaluate sequence predictions we use two metrics based on the Levenshtein distance: edit score (ES) and action error rate (AER) (inspired by the word-error rate metric used in speech recognition). The Levenshtein distance, $\operatorname{L}(G,P)$, is the minimum number of insertions, deletions, and substitutions required to convert a predicted sequence $P$ to a ground-truth sequence $G$. For example, if $G$ = [\emph{reach}, \emph{idle}, \emph{stabilize}] and $P$ = [\emph{reach}, \emph{transport}], then $\operatorname{L}(G,P)=2$ (\emph{transport} is substituted for \emph{idle} and \emph{stabilize} is inserted). We have:
\begin{equation}
    \operatorname{ES}(G,P) := \left(1 - \frac{\operatorname{L}(G,P)}{\max(\operatorname{len}(G),\operatorname{len}(P))}\right) \times 100, \qquad \operatorname{AER}(G,P):=\frac{\operatorname{L}(G,P)}{\operatorname{len}(G)}
\end{equation} 
where $\operatorname{len}(G)$ and $\operatorname{len}(P)$ are the lengths of the ground-truth and predicted sequence respectively. 

The edit score is more lenient when the estimated sequence is longer. In contrast, AER penalizes longer and shorter predictions equally. For example, if $G$ = [\emph{reach}, \emph{idle}, \emph{stabilize}], $P_1$ = [\emph{reach},\emph{idle}], and $P_2$ = [\emph{reach}, \emph{idle}, \emph{stabilize}, \emph{transport}], then $\operatorname{ES}(G,P_1)$ = 0.67 and $\operatorname{ES}(G,P_2)$ = 0.75, but $\operatorname{AER}(G,P_1)$ = $\operatorname{AER}(G,P_2)$ = 0.33. 
 
\begin{figure}[t]
\begin{minipage}{0.49\textwidth}
\begin{center}
    \includegraphics[width=0.9\textwidth]{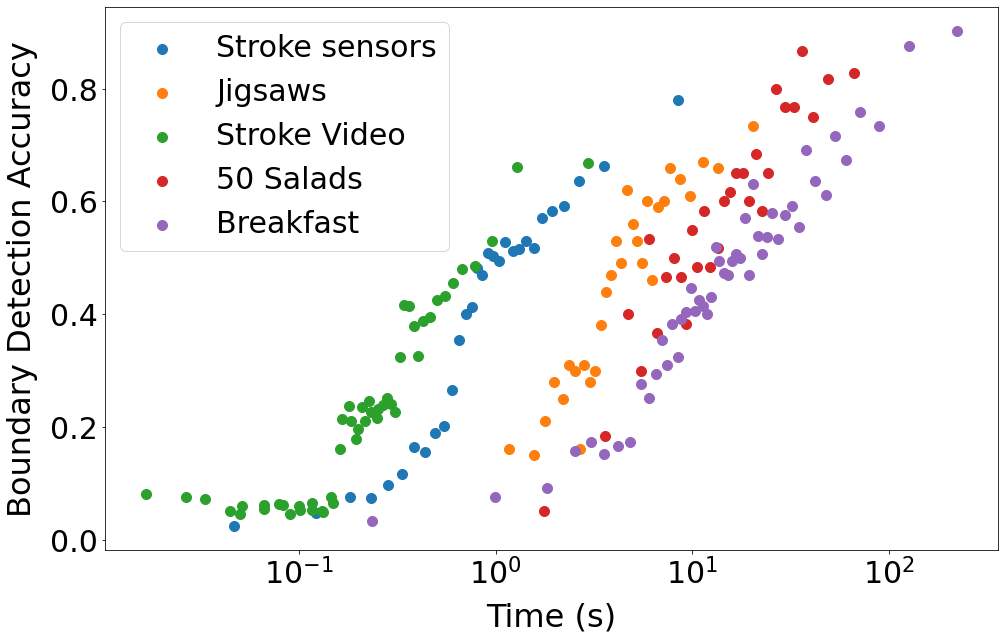}
  \end{center}
  \caption{Boundary accuracy achieved by the segmentation models vs duration of the actions for several datasets. Boundary-detection accuracy is inversely proportional to action duration.}
\label{fig:boundary_time}
\end{minipage}\hfill
\begin{minipage}{0.49\textwidth}
\begin{center}
    \includegraphics[width=\textwidth]{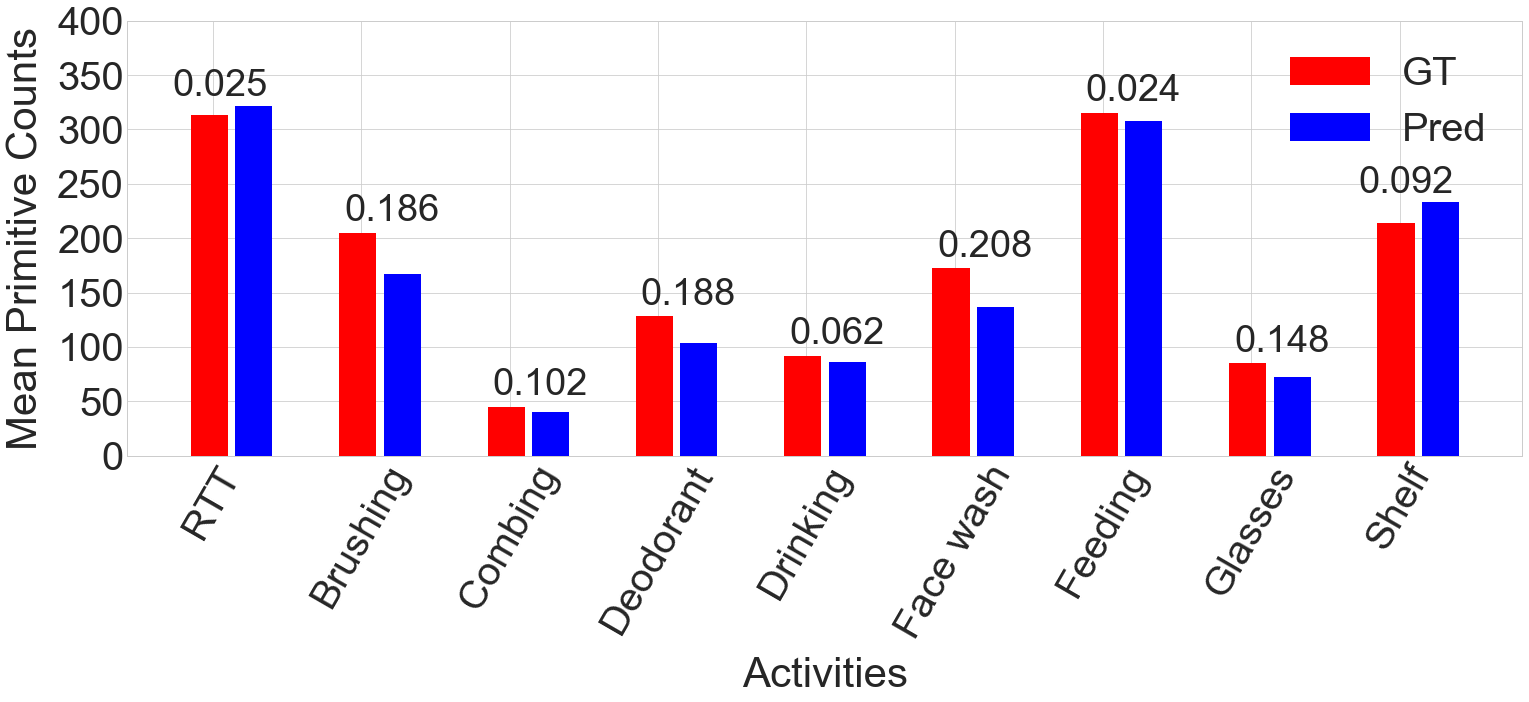}
  \end{center}
  \caption{Comparison of ground-truth and predicted mean counts for the different activities in the StrokeRehab dataset. The relative error is very small for structured activities like moving objects on/off a shelf (Shelf), and larger for unstructured activities like brushing.}
\label{fig:primi_count}
\end{minipage}
\end{figure}


\subsection{Implementation Details of the Sequence-to-Sequence (Seq2seq) model} \label{subsec:seq2seq imple}



During training we apply the seq2seq model on overlapping windows of short duration. During inference, we divide the input data into non-overlapping windows and concatenate the estimates removing duplicates at the boundaries. 
We implement two variants of the seq2seq method:

\begin{itemize}[leftmargin=*]
    \item \emph{Raw2seq} uses raw sequences of sensor data or video features as input.
    
    
    \item \emph{Seg2seq} uses frame-wise predictions from a segmentation-based model as inputs (specifically the MS-TCN baseline model in Section~\ref{subsec:seg_models}). The motivation is that the \emph{raw2seq} has a limited field-of-view, but
    identifying high-level actions requires modeling long-term dependencies, which can be captured more easily by segmentation models. 
\end{itemize}
Additional implementation details are provided in Appendix~\ref{app_subsec:model_imple}

\subsection{Segmentation Models} \label{subsec:seg_models}

To ensure a fair comparison, we optimized our selected segmentation models using our metric of interest for segmentation prediction (AER). The evaluation procedure is explained in detail in  Section~\ref{app_subsec:evaluation}.

\textbf{Baseline}: We use the MS-TCN~\citep{farha2019ms} model with an improved loss function suggested in ASRF~\citep{ishikawa2021alleviating} as a baseline model. 
 
\textbf{Baseline + boundary detection:} This is the ASRF  method~\citep{ishikawa2021alleviating}, a state-of-the-art model for action segmentation. It refines the prediction of the baseline model using boundary predictions. 

\textbf{Baseline + smoothing window:} We use a smoothing window to improve the baseline prediction. 

\subsection{Results}
\label{sec:seq2seq_model}


Table~\ref{tab:result_stroke_video} and Table~\ref{tab:result_other_datasets} show that sequence-to-sequence models outperform segmentation-based models on all the datasets.
The raw2seq version achieves better performance on the sensor data of the StrokeRehab data, where the actions are very localized and do not require modeling long time dependencies, whereas the seg2seq version is superior on the remaining datasets. The baselines achieve edit score values that are close to those of seq2seq on 
the sensor data of StrokeRehab (but not on the video data), and on 50Salads, but the AER of seq2seq is better (as explained in Section~\ref{sec:metrics} the edit score is more lenient with the false positives that tend to be produced by segmentation-based models). Interestingly, the refinement strategies of smoothing and boundary detection do not improve the baseline on the StrokeRehab, highlighting that it contains actions that are qualitatively different from those of the other datasets.



\begin{table}[t]
\centering
\resizebox{1\linewidth}{!}{
\begin{tabular}{>{\centering\arraybackslash}m{0.15\linewidth} >{\centering\arraybackslash}m{0.30\linewidth}| >{\centering\arraybackslash}m{0.20\linewidth} >{\centering\arraybackslash}m{0.22\linewidth}| >{\centering\arraybackslash}m{0.22\linewidth}>{\centering\arraybackslash}m{0.22\linewidth}}
\toprule
   & \multirow{2}{*}{Model} &  \multicolumn{2}{c|}{Video Data} & \multicolumn{2}{c}{Sensor Data}  \\
   \cline{3-6}
   & & {Edit Score} & {Action Error Rate} &  {Edit Score} & {Action Error Rate}\\
\midrule[0.7pt]
& {Baseline}  & 62.2 (60.8 - 63.6) & 0.392 (0.371 - 0.413) & 68.9 (67.3 - 70.5) & 0.330 (0.307 - 0.354)   \\
Segmentation & {+ Boundary detection} & {58.7 (57.3 - 60.2)} & 0.436 (0.417 - 0.456) &  67.9 (66.3 - 69.5) & 0.349 (0.326 - 0.372)  \\
model &{ + Smoothing window }  & 62.7 (61.3 - 64.1) & 0.390 (0.370, 0.410) &  68.8 (67.3 - 70.3) & 0.317 (0.297 - 0.338)\\
\midrule[0.7pt]
\multirow{2}{*}{Seq2seq}&{Seg2seq} & \textbf{{67.6 (66.4 - 68.8)}}& \textbf{{0.322 (0.307 - 0.339)}}  & 63.0 (61.3 - 64.7) & 0.337 (0.311 - 0.363) \\
&{Raw2seq } & {66.6 (65.4 - 67.9)} & {0.329 (0.312 - 0.345)} &\textbf{69.0 (67.3 - 70.6)} & \textbf{0.308 (0.287 - 0.329)}  \\
\bottomrule
\end{tabular}
}
\caption{Results on \textit{StrokeRehab}: Seq2seq outperforms segmentation-based approaches. We report mean (95\% confidence interval) which is computed via bootstrapping (see Appendix~\ref{app_subsec:evaluation}). 
}
    \label{tab:result_stroke_video}
\end{table}

\begin{table}[t]
\centering
\resizebox{1\linewidth}{!}{
\begin{tabular}{>{\centering\arraybackslash}m{0.15\linewidth} >{\centering\arraybackslash}m{0.25\linewidth}| >{\centering\arraybackslash}m{0.10\linewidth} >{\centering\arraybackslash}m{0.15\linewidth}| >{\centering\arraybackslash}m{0.10\linewidth}>{\centering\arraybackslash}m{0.15\linewidth}|>{\centering\arraybackslash}m{0.10\linewidth}>{\centering\arraybackslash}m{0.15\linewidth}}
\toprule
   & \multirow{2}{*}{Model} &  \multicolumn{2}{c|}{50 salads} & \multicolumn{2}{c|}{Breakfast} & \multicolumn{2}{c}{Jigsaws} \\
   \cline{3-8}
   & & {Edit Score} & {Action Error Rate} &  {Edit Score} & {Action Error Rate} & {Edit Score} & {Action Error Rate}\\
\midrule[0.7pt]
& {Baseline}  & 70.8 & 0.43 &  61.7 & 0.97 & 61.44  & 0.82 \\
Segmentation & {+ Boundary detection} & 75.2 & 0.33 &  70.9  & 0.45 &  74.63   & 0.31 \\
model &{ + Smoothing window }  & 76.4 & 0.32 &  69.1  & 0.51 & 76.54 & 0.31 \\
\midrule[0.7pt]
\multirow{2}{*}{Seq2seq}&{Seg2seq} & \textbf{76.9} & \textbf{0.30} &   \textbf{73.7} &  \textbf{0.37} & \textbf{83.87 } &  \textbf{0.17 }   \\
&{Raw2seq } &  69.4 & 0.54 &  64.1 & 0.55 & 70.13 &0.35\\
\bottomrule
\end{tabular}
}
\caption{Results on action-recognition benchmarks: Seg2seq, the seq2seq model which uses the output of a pretrained segmentation-based model, outperforms segmentation-based approaches.
}
    \label{tab:result_other_datasets}
\end{table}
\vspace{-3mm}

\subsection{Application: Data-driven Stroke Rehabilitation} \label{subsec:application_rehab_quant}
In stroke rehabilitation, action identification can be used for quantifying  dose by counting functional primitives.
Figure~\ref{fig:primi_count} show that the raw2seq version of the seq2seq model produces accurate counts for all activities in the StrokeRehab dataset. Performance is particularly good for structured activities such as moving objects on/off a shelf, in comparison to less structured activities such as brushing, which tend to be more heterogeneous across patients (detailed count-based results are provided in Appendix~\ref{app_sec:tpr_fdr} and Appendix~\ref{app_sec:count_pred}). Figure~\ref{fig:confusion_matrix_video} and Figure~\ref{fig:confusion_matrix_sensor} show confusion matrices of the predictions produced by the best models based on video and sensor data respectively. The sensor-based model has difficulties differentiating between \emph{idle} and \emph{transport}, whereas the video-based model confuses \emph{idle} and \emph{reach}. 

\section{Discussion and Conclusion}
\vspace{-3mm}
In this work, we introduce a large-scale, multimodal dataset, StrokeRehab, as a new benchmark for identification of action sequences that includes subtle short-duration actions labeled at a high temporal resolution. In addition, we introduce a novel sequence-to-sequence approach, which outperforms existing methods on StrokeRehab, as well as on existing benchmark datasets. A known limitation of sequence-to-sequence approaches is that they have difficulties capturing long-term dependencies. Here, we address this by using the output of a segmentation-based network as an input for the sequence-to-sequence model. An interesting direction for future research is to design sequence-to-sequence models capable of directly learning these dependencies. Our results also show that models based on video and wearable-sensor data have different strengths and weaknesses (see Section~\ref{subsec:application_rehab_quant}), which suggests that multimodal approaches may have significant potential.

\subsubsection*{Acknowledgments}
We would like to thank the volunteers who contributed to label the dataset: Ronak Trivedi, Adisa Velovic, Sanya Rastogi, Candace Cameron, Sirajul Islam, Bria Bartsch, Courtney Nilson, Vivian Zhang, Nicole Rezak, Christopher Yoon, Sindhu Avuthu, and Tiffany Rivera. We thank Dawn Nilsen, and OT EdD for expert advice on the testing battery. This work was supported by an AHA postdoctoral fellowship 19AMTG35210398 (AP), NIH grants R01 LM013316 (AK, KL, HR, CFG, HMS) and K02 NS104207 (HMS), NSF NRT-HDR Award 1922658 (AK, KL, HR, CFG).

\newpage
\bibliography{iclr2022_conference}
\bibliographystyle{iclr2022_conference}
\newpage
\appendix

\section{Confusion matrices for the StrokeRehab dataset}
\begin{figure}[h]
\begin{minipage}{0.49\textwidth}
\centering
\includegraphics[width = \textwidth]{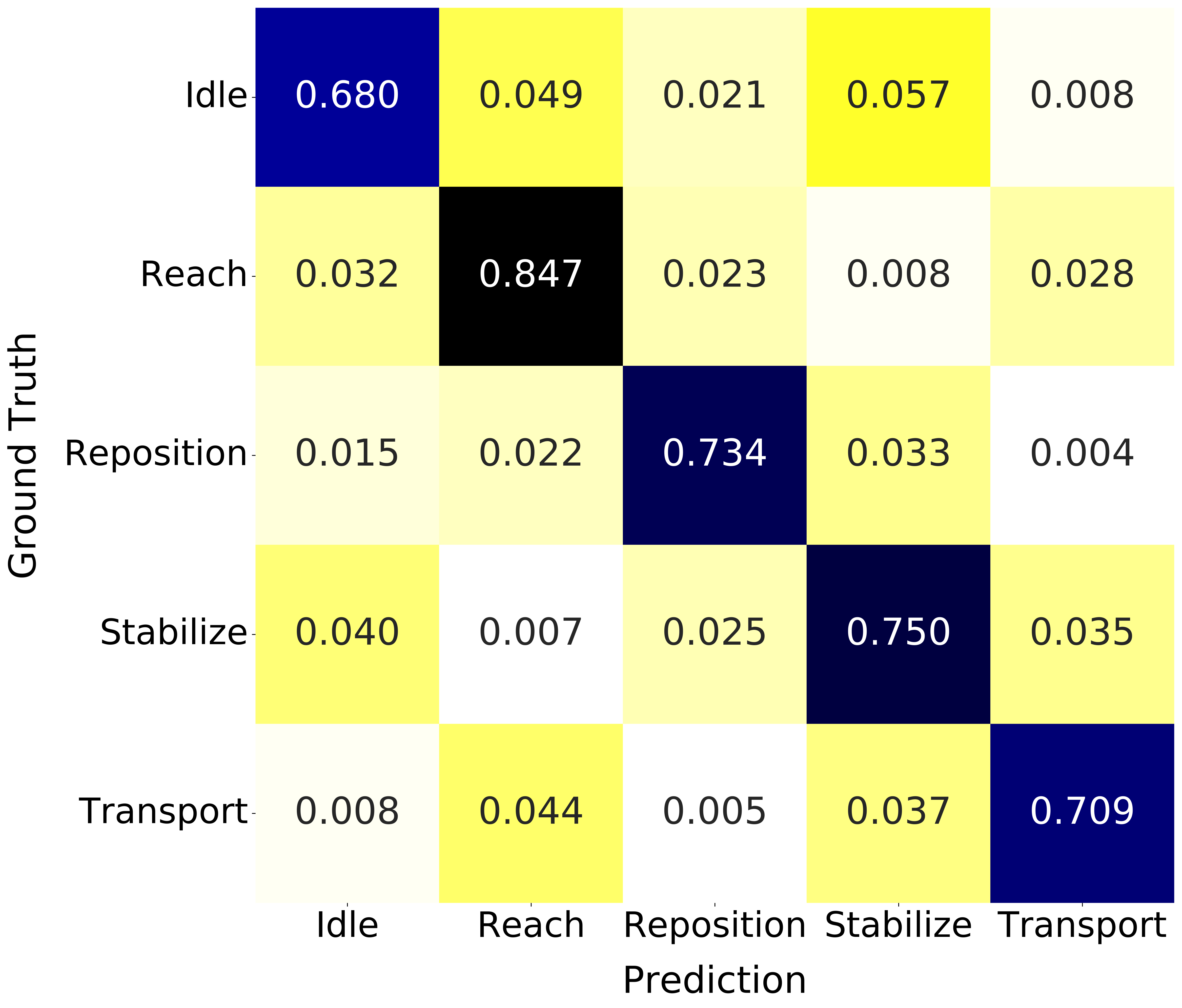}
\caption{Confusion Matrix for the best performing model on the StrokeRehab video Dataset. The model makes the most mistakes for the \emph{idle} primitive. }
\label{fig:confusion_matrix_video}
\end{minipage}\hfill
\begin{minipage}{0.49\textwidth}
\centering \includegraphics[width=\textwidth]{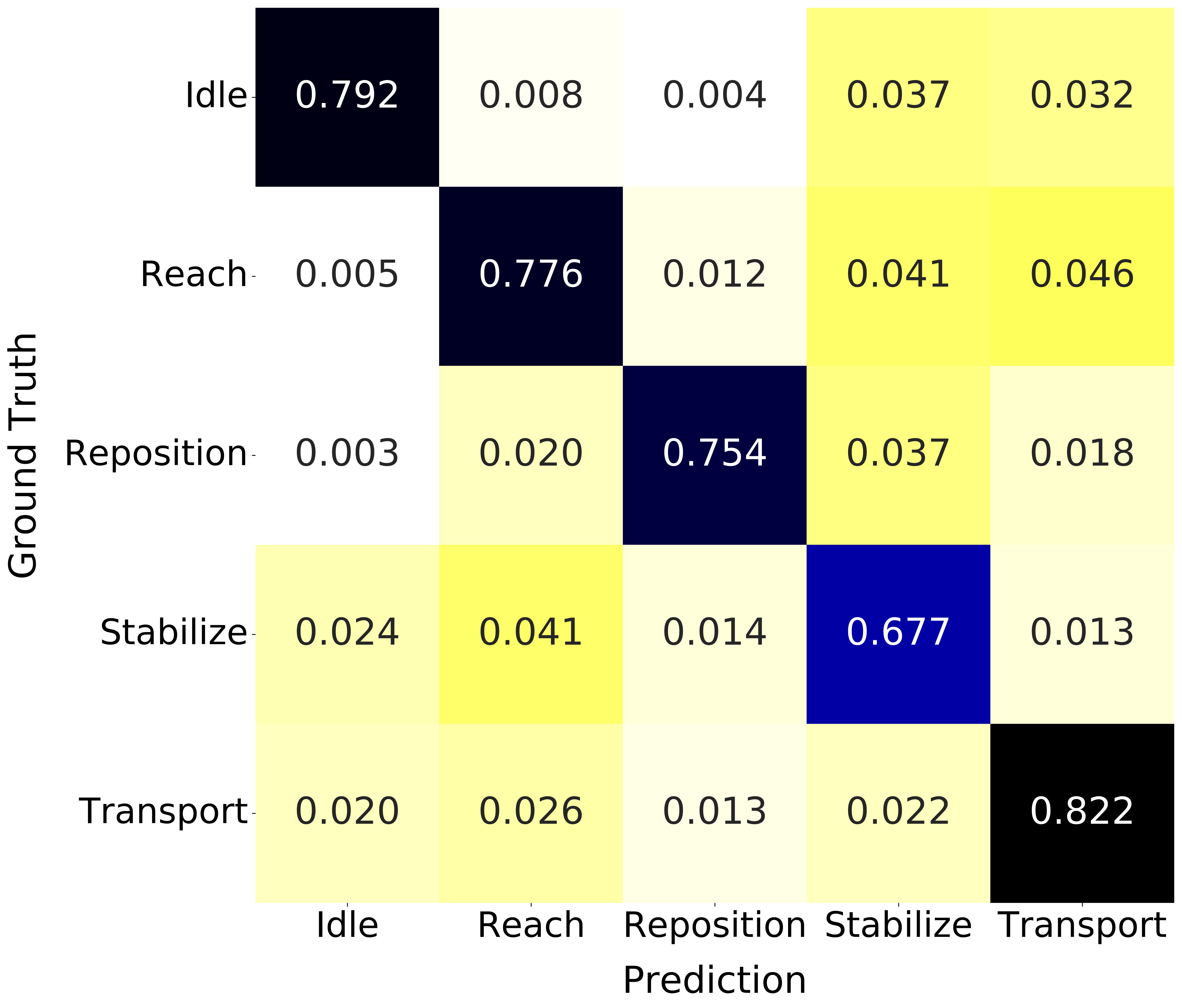}
\caption{Confusion Matrix for the best performing model on the StrokeRehab sensor Dataset. The model makes the most mistakes for the \emph{stabilize} primitive. }
\label{fig:confusion_matrix_sensor}
\end{minipage}
\end{figure}

Figure~\ref{fig:confusion_matrix_video} and Figure~\ref{fig:confusion_matrix_sensor} show confusion matrices for the best performing model on the stroke video and the stroke sensor dataset (\emph{Seg2seq} and \emph{Raw2Seq} respectively).
The confusion matrices are built using the total number of substitutions and correct predictions from both models. Specifically, each diagonal entry of the matrix represents the fraction of ground-truth primitives that were estimated correctly. Each off-diagonal entry represents the fraction of ground-truth primitives that were substituted.  We found that the nature of errors of two models were complementary to some extent. For example, the model trained on video data confused reaches with idles (Figure~\ref{fig:confusion_matrix_video}) whereas the model trained on sensor data confused reaches with stabilizes 
(Figure~\ref{fig:confusion_matrix_sensor}). For the video data, confusion occurs mainly between primitives which are usually performed one after the other (eg. idles and reaches). In contrast, for the sensor data, there is more confusion between primitives that look similar, such as reaches and transports or idles and stabilizes. 

\section{False Discovery Rate and True Positive rate} \label{app_sec:tpr_fdr}

In addition to edit score and action error rate, we also evaluated the model performance using the true positive rate (TPR) and the false discovery rate (FDR). To compute these quantities we compared the estimated sequence to the ground truth actions to determine the number of actions that are correctly identified, incorrectly identified (e.g. the model estimates a transport, but the ground truth is a reach), missed (an action present in the ground truth is not present in the estimate), or spurious (an action present in the estimate is not present in the ground truth). The TPR is the ratio between the correctly-identified actions and the total ground-truth actions. The FDR is the ratio between the wrong predictions (incorrectly identified and spurious) and the total estimated actions. It should be noted that there is a trade-off between TPR and FDR. Therefore, to compare the various models, both TPR and FDR can be combined using $\text{F1-score}$, which is the harmonic mean of TPR and 1-FDR: $ \operatorname{F1} :=\frac{ 2(1-\operatorname{FDR})\operatorname{TPR}}{1-\operatorname{FDR}+\operatorname{TPR}}$.

\begin{table}[ht]
\centering
\resizebox{1\linewidth}{!}{
\begin{tabular}{>{\centering\arraybackslash}m{0.15\linewidth} >{\centering\arraybackslash}m{0.30\linewidth}| >{\centering\arraybackslash}m{0.22\linewidth} >{\centering\arraybackslash}m{0.22\linewidth}| >{\centering\arraybackslash}m{0.22\linewidth}>{\centering\arraybackslash}m{0.22\linewidth}}
\toprule
   & \multirow{2}{*}{Model} &  \multicolumn{2}{c|}{Video Data} & \multicolumn{2}{c}{Sensor Data}  \\
   \cline{3-6}
   & & {FDR} & {TPR} &  {FDR} & {TPR}\\
\midrule[0.7pt]
& {Baseline}  & 0.148 (0.135 - 0.161) & 0.643 (0.622- 0.664) & 0.201 (0.186 - 0.216) & {0.790 (0.770 - 0.812)}  \\
Segmentation & {+ Boundary detection} & {0.126 (0.113 - 0.139)} & 0.585 (0.566 - 0.605) &   0.219 (0.205 - 0.233) &  0.784 (0.763 - 0.803)  \\
model &{ + Smoothing window }  & 0.180 (0.166 - 0.194) &0.666 (0.646 - 0.685) & 0.162 (0.149 - 0.173) & 0.758 (0.737 - 0.779) \\
\midrule[0.7pt]
\multirow{2}{*}{Seq2seq}&{Seg2seq} & 0.211 (0.200 - 0.221) &{0.743} (0.732 - 0.753)& {0.145 (0.134 - 0.157)} & 0.707 (0.678 - 0.734)  \\
&{Raw2seq } & 0.216 (0.207 - 0.226) &0.734 (0.722 - 0.744) & 0.166 (0.153 - 0.179) & 0.767 (0.747 - 0.786) \\
\bottomrule
\end{tabular}
}
\caption{Comparison of seq2seq model and action segmentation model on \textit{StrokeRehab}: in terms of FDR and TPR.
}
    \label{tab:result_stroke_video_FDRTPR}
\end{table}

\begin{table}[t]
\centering
\resizebox{1\linewidth}{!}{
\begin{tabular}{>{\centering\arraybackslash}m{0.15\linewidth} >{\centering\arraybackslash}m{0.25\linewidth}| >{\centering\arraybackslash}m{0.10\linewidth} >{\centering\arraybackslash}m{0.15\linewidth}| >{\centering\arraybackslash}m{0.10\linewidth}>{\centering\arraybackslash}m{0.15\linewidth}|>{\centering\arraybackslash}m{0.10\linewidth}>{\centering\arraybackslash}m{0.15\linewidth}}
\toprule
   & \multirow{2}{*}{Model} &  \multicolumn{2}{c|}{50 salads} & \multicolumn{2}{c|}{Breakfast} & \multicolumn{2}{c}{Jigsaws} \\
   \cline{3-8}
   & & {FDR} & TPR &  {FDR} & {TPR} & {FDR} & {TPR}\\
\midrule[0.7pt]
& {Baseline}  & 0.26 & 0.86 & 0.34 & 0.82 & 0.357 & 0.898 \\
Segmentation & {+ Boundary detection} & 0.17 & 0.80 & 0.19 & 0.77 & 0.191 & 0.846 \\
model &{ + Smoothing window }  & 0.18 & 0.83 & 0.23 & 0.79 & 0.174 & 0.878 \\
\midrule[0.7pt]
\multirow{2}{*}{Seq2seq}&{Seg2seq} & 0.17 & 0.82 &0.17&0.79 & 0.092 & 0.874  \\
&{Raw2seq } & 0.31 & 0.76 & 0.28 & 0.70 &0.218 &0.780  \\
\bottomrule
\end{tabular}
}
\caption{Comparison of seq2seq model and action segmentation model on 50Salads, Breakfast, Jigsaws: in terms of FDR and TPR.
}
    \label{tab:result_other_datasets_FDRTPR}
\end{table}

\section{Count Prediction} \label{app_sec:count_pred}
\begin{table}[H]
\centering
\resizebox{1\linewidth}{!}{
\begin{tabular}{|>{\centering\arraybackslash}m{0.15\linewidth} |  >{\centering\arraybackslash}m{0.1\linewidth} >{\centering\arraybackslash}m{0.1\linewidth} >{\centering\arraybackslash}m{0.1\linewidth} >{\centering\arraybackslash}m{0.1\linewidth}>{\centering\arraybackslash}m{0.1\linewidth}| >{\centering\arraybackslash}m{0.1\linewidth} >{\centering\arraybackslash}m{0.1\linewidth} >{\centering\arraybackslash}m{0.1\linewidth} >{\centering\arraybackslash}m{0.1\linewidth}>{\centering\arraybackslash}m{0.1\linewidth}|
}
\toprule 
\multirow{2}{*}{Subject} &  \multicolumn{5}{c|}{Ground-Truth}&  \multicolumn{5}{c|}{Prediction}  \\
   \cline{2-11}
   & {\emph{idle}} & {\emph{reach}} &  {\emph{reposition}} & {\emph{transport}} & {\emph{stabilize}} & {\emph{idle}} & {\emph{reach}} &  {\emph{reposition}} & {\emph{transport}} & {\emph{stabilize}} \\
\midrule[0.7pt]
s4 & 397&	194&	290&	415&	301 & 293 &	213 &	204 &	370 &	281 \\
 s17 & 282&	353	&257&	489&	567 & 252 &	348 &	225 &	582	& 724\\
 s26  & 237&	293&	211&	245&	319 &   240 &	253 &	211 &	183 &	326\\
 s37  & 188&	337&	191&	266&	464 &  202 &	328 &	174	& 113 &	408\\
 s39  & 338&	273&	250&	228&	294 & 324 &	252 &	219 &	389 &	366\\
 s42  & 290&	328&	245&	330&	470 & 287 &	297 &	228 &	292 &	417 \\
 s44  & 298 & 365 &	264	& 294	& 429 & 285 &	275 &	229 &	327 &	406 \\
 s47  & 215 & 313 &	203 &	186	 &427 & 181 &	287 &	156 &	74 &	332 \\
 \midrule
Mean  &  280.62 &	307	& 238.87 &	306.62 &	408.87 & 258 &	281.62 &	205.75 &	291.25 &	407.5\\
Std-Dev  & 63.26 &	51.21 &	31.8 &	94.8 &	90.43 & 45.52 &	40.73  & 25.24 &	155.02 &	127.64\\

\bottomrule
\end{tabular}
}
\caption{Number of ground-truth functional primitives for each subject in the test set of the StrokeRehab sensor data, compared to the number predicted by the best performing model. }
    \label{tab:stroke_sensor_conf_results}
\end{table}

\begin{table}[H]
\centering
\resizebox{1\linewidth}{!}{
\begin{tabular}{|>{\centering\arraybackslash}m{0.15\linewidth} |  >{\centering\arraybackslash}m{0.1\linewidth} >{\centering\arraybackslash}m{0.1\linewidth} >{\centering\arraybackslash}m{0.1\linewidth} >{\centering\arraybackslash}m{0.1\linewidth}>{\centering\arraybackslash}m{0.1\linewidth}| >{\centering\arraybackslash}m{0.1\linewidth} >{\centering\arraybackslash}m{0.1\linewidth} >{\centering\arraybackslash}m{0.1\linewidth} >{\centering\arraybackslash}m{0.1\linewidth}>{\centering\arraybackslash}m{0.1\linewidth}|
}
\toprule 
\multirow{2}{*}{Subject} &  \multicolumn{5}{c|}{Ground-Truth}&  \multicolumn{5}{c|}{Prediction}  \\
   \cline{2-11}
   & {\emph{idle}} & {\emph{reach}} &  {\emph{reposition}} & {\emph{transport}} & {\emph{stabilize}} & {\emph{idle}} & {\emph{reach}} &  {\emph{reposition}} & {\emph{transport}} & {\emph{stabilize}} \\
\midrule[0.7pt]
s4 & 151 & 282 & 229 & 363 & 307 & 153 & 402 & 123 & 416 & 160 \\
 s17 & 336 & 541 & 224 & 466 & 261 & 263 & 634 & 231 & 493 & 266\\
 s26  &293 & 315 & 202 & 235 & 221 & 247 & 320 & 195 & 208 & 199\\
 s37  & 251 & 356 & 95 & 219 & 90 & 201 & 369 & 84 & 239 & 78\\
 s39  & 142 & 165 & 128 & 182 & 205 & 119 & 199 & 125 & 212 & 162\\
 s42  &219 & 351 & 136 & 291 & 157 & 188 & 309 & 120 & 258 & 156\\
 s44  & 223 & 282 & 143 & 193 & 160 & 151 & 227 & 147 & 155 & 156 \\
 s47  & 198 & 304 & 98 & 174 & 79 & 162 & 242 & 99 & 181 & 83\\
 \midrule
Mean  &  226.62 & 324.5 & 156.87 & 265.37 & 185 & 185.5 & 337.75 & 140.5 & 270.25 & 157.5\\
Std-Dev  & 61.98 & 98.82 & 50.62 & 96.21 & 74.18 & 46.44 & 129.76 & 46.21 & 112.08 & 56.34\\

\bottomrule
\end{tabular}
}
\caption{Number of ground-truth functional primitives for each subject in the test set of the StrokeRehab video data, compared to the number predicted by the best performing model.}
    \label{tab:stroke_video_conf_results}
\end{table}

Tables~\ref{tab:stroke_sensor_conf_results} and~\ref{tab:stroke_video_conf_results}, show the number of ground-truth functional primitives for each subject in the test set of the StrokeRehab, compared to the number predicted by the best performing model. 
We observe that the predicted counts of the functional primitives are consistently close to their corresponding ground truth counts. The number of ground-truth actions for the sensor and video data because some labels were filtered out during preprocessing. 

\section{Experimental details}

\subsection{Benchmark datasets} \label{app_subsec:datasets}
\paragraph{50 salads dataset}

50 salads dataset~\citep{stein2013combining} contains 50 videos with 17 action classes. In this dataset, 25 people in total prepare two kinds of mixed salads. Each video contains 9000 to 19000 RGB frames. On average, each video contains 20 action instances and is 6.4 minutes long. For evaluation, we follow~\cite{ishikawa2021alleviating}, performing five-fold cross-validation and reporting the average. As per \cite{farha2019ms}, we use I3D model~\citep{carreira2017quo}, pretrained on the Kinetics dataset, to extract spatio-temporal features from the videos and then use them as the input for our models. 

\paragraph{Breakfast dataset}

Breakfast dataset~\citep{kuehne2014language} contains 1712 videos, which are recorded in 18 different kitchens displaying activities of breakfast preparation. There are 48 different actions. On average, each video contains 6 action instances.  For evaluation, we follow~\cite{ishikawa2021alleviating}, performing four-fold cross-validation and reporting the average. As per \cite{farha2019ms}, we use I3D model~\citep{carreira2017quo}, pretrained on the Kinetics dataset, to extract spatio-temporal features from the videos and then use them as the input for our models.  

\paragraph{Jigsaws dataset}

Jigsaws dataset~\citep{gao2014jhu} contains 103 surgical activities of 3 types - Knot tying, Suturing and Needle Passing performed by 8 subjects using a robotic surgical system. The subjects have varying degrees of robotic surgical experience. The kinematic data from the robotic arm and the videos of the activities are available. The activities contain 14 different actions overall. On average, each video contains 17 action instances. We split the subjects into 4 folds,where each fold contains two subjects with different levels of robotic experience.  For evaluation, we follow~\cite{ishikawa2021alleviating}, performing four-fold cross-validation and reporting the average.

\subsection{Evaluation} \label{app_subsec:evaluation}

In the case of the action segmentation models, 
we use a small validation set separate from the training set of each fold to select models based on Action Error Rate (AER), and test the performance on the validation set of that fold. We then follow the same evaluation as previous works~\citep{farha2019ms,ishikawa2021alleviating} by simply averaging the result for all the folds. For StrokeRehab, there is a held-out test set.  To perform model selection, we apply 4-fold cross-validation. For each fold, we select the best model according to the best Action Error Rate (AER) on the validation set. We then ensemble the prediction of the models trained on 4 folds and evaluate it on the held-out test set. For the action segmentation models, we perform ensembling by averaging the model outputs. In the case of the seq2seq model, the decoding of each action depends on the previous prediction. 
To ensemble, average the output of the models at each step and use this average prediction as the previous prediction in the decoder for each individual model.
 
 \textbf{Confidence interval computation}: We report the 95\% confidence interval for the StrokeRehab dataset by creating 1,000 bootstrap replicates of the test set. We calculate the upper and lower confidence limit using $\mu \pm 1.96\sigma/\sqrt{N}$ where $\mu$ and $\sigma$ are the mean and standard deviation of the performance measure of interest, and $N$ the is number of bootstrap replicates.

\subsection{Implementation of models} \label{app_subsec:model_imple}
\subsubsection{Action Segment Refinement Framework (ASRF)}
The ASRF model has two modules: one for frame-wise action segmentation and another for boundary detection. 
The backbone of both modules is a MS-TCN model with several convolutional stages, each composed of multiple layers of dilated residual convolutions.
To implement our Baseline model (see Section~\ref{subsec:seg_models}), we just use the segmentation module. To implement our Baseline + Boundary Detection model (see Section~\ref{subsec:seg_models}), we refine the output of the segmentation module using the boundaries detected by the boundary detection module.

The loss function used to train this model is combination of two loss functions: weight-cross entropy for frame-wise action classification and boundary detection. We use a regularization parameter $\lambda$ to determine the weight of the boundary-detection term in the loss.

Below we provide some specific implementation details for each specific dataset.

\textbf{StrokeRehab}: We use a backbone MS-TCN model with 4 convolutional stages where each stage has 10 layers of dilated residual convolutions, outputting 64 channels. 
For the sensor data, the parameter $\lambda$ in the loss function was set to 0.1. For the video data, $\lambda$ was set to 1. These values were determined based on preliminary cross-validation experiments on the training data.

\textbf{50Salads and Breakfast}: We follow the settings described in the original paper~\citep{ishikawa2021alleviating}. In particular, we use the same backbone MS-TCN model as for the StrokeRehab dataset. The only difference is that we perform model selection based on the best validation AER rather than the frame-wise accuracy. We do this to achieve a fair comparison with the seq2seq models, since AER is our metric of interest.

\textbf{Jigsaws}: 
We use a backbone MS-TCN model with 4 convolutional stages where each stage has 15 layers of dilated residual convolutions, outputting 128 channels.
The $\lambda$ parameter for the boundary detection loss is set to 0.1, based on preliminary cross-validation experiments on the training data.

\subsubsection{Sequence-to-sequence (Seq2seq) }
All seq2seq models are trained with the following hyper-parameters: learning rate = 5e-4, dropout = 0.1, weight decay = 0.0001, num of epochs = 150. When dividing the whole sequence into time windows, we zero pad the last window. 

\textbf{StrokeRehab video dataset}: 

The encoder module is an MS-TCN model with 4 convolutional stages where each stage has 15 layers of dilated residual convolutions, outputting 256 channels. The decoder module is an LSTM with attention mechanism and a 512-dimensional hidden representation.

\begin{itemize}[leftmargin=*]
    \item \emph{Raw2seq}. During training, we divide each video sequence of 432 dimensional raw feature vectors into 240-frame windows with an overlap of 100 frames. We then train the model using the windows. During inference, we divide each testing video sequence into non-overlapping 240-frame windows and input that to the model. We then concatenate the resulting estimates from each non-overlapping window and remove the duplicates. 
    \item \emph{Seg2seq}. We use the baseline segmentation model to obtain frame-wise prediction probabilities. We then input the frame-wise probabilities to the seq2seq model. During training, we divide the softmax probabilities of each video sequence into 240-frame windows with an overlap of 100 frames and train the model using these windowed inputs. During inference, we divide each testing video sequence into non-overlapping 240-frame windows and input them to the model. We then concatenate the result from the non-overlapping window and remove the duplicates.
\end{itemize}

\textbf{StrokeRehab sensor dataset}: 
In order to apply the seq2seq models to the sensor data, we window the input sequence. During training, we use overlapping six-second windows, where the labels only correspond to the middle four seconds. During inference, we use non-overlapping windows and concatenate the result, removing duplicates. In addition to the seq2seq cost function, we also incorporate a segmentation frame-wise loss on the output of the encoder.
\begin{itemize}[leftmargin=*]
    \item \emph{Raw2seq}. The input to the model are 77-dimensional sensor measurements. The encoder module is a three layered bi-LSTM model with a 3072 dimensional hidden representation. The decoder module is also an LSTM with a 6144 dimensional hidden representation. 
    \item \emph{Seg2seq}. We use the baseline segmentation model to obtain frame-wise prediction probabilities. We then input the frame-wise probabilities to the seq2seq model. The encoder module is a three layered bi-LSTM model with 256 dimensional hidden representation. The decoder module is also an LSTM with 512 dimensional hidden representation. 
\end{itemize}

\textbf{50Salads}: The encoder module is an MS-TCN model with 4 convolutional stages where each stage has 15 layers of dilated residual convolutions, outputting 256 channels. The decoder module is an LSTM with attention mechanism and a 512-dimensional hidden representation.

\begin{itemize}[leftmargin=*]
    \item \emph{Raw2seq}. During training, we divide each video sequence of 1600 dimensional raw feature vectors to 500-frame windows with overlap of 100 frames. 
    During inference, we divide each testing video sequence into non-overlapping 1600-frame windows. We then concatenate the result and remove duplicates.
    \item \emph{Seg2seq}. We use the baseline segmentation model to obtain frame-wise prediction probabilities. We then input the frame-wise probabilities to the seq2seq model. During training, we divide the probabilities of each video sequence into 450-frame windows with an overlap of 100 frames. We then train the model using the windowed data. During inference, we divide each testing video sequence into  non-overlapping 450-frame windows. We then concatenate the result and remove  duplicates.
    
\end{itemize}

\textbf{Breakfast}: The encoder module is an MS-TCN model with 4 convolutional stages where each stage has 15 layers of dilated residual convolutions, outputting 256 channels. The decoder module is an LSTM with attention mechanism and a 512-dimensional hidden representation.

\begin{itemize}[leftmargin=*]
    \item \emph{Raw2seq}. During training, we divide each video sequence of 1600 dimensional raw feature vectors into 500-frame windows with an overlap of 100 frames. During inference, we divide each testing video sequence to  non-overlapping 1600-frame windows. We then concatenate the result and remove  duplicates.
    \item \emph{Seg2seq}. We use the baseline segmentation model to obtain frame-wise prediction probabilities. We then input the frame-wise probabilities to the seq2seq model. During training, we divide the probabilities of each video sequence into 800-frame windows with an overlap of 200 frames. During inference, we divide each testing video sequence into  non-overlapping 800-frame windows. We then concatenate the result and remove duplicates. 
\end{itemize}

\textbf{Jigsaws}: 

\begin{itemize}[leftmargin=*]
    \item \emph{Raw2seq}. Seq2seq for raw features of Jigsaws has an encoder, a decoder module and attention mechanism. The encoder is a MS-TCN model with 4 stages of convolutions with each stage having 10 layers of dilated residual convolutions outputting 128 channels. The decoder is a GRU-RNN with 256 dimensional hidden representation. We also used a multi-headed attention mechanism (2 heads) with a multi-layer perceptron producing 128 dimensional representation. During training, we divide each Kinematic input sequence of 38 dimensional raw feature vectors to 400-frame windows with an overlap of 350 frames. We then train the model using these windows. During inference, we divide each testing video sequence to  non-overlapping 400-frame windows and input that to the model. We then concatenate the result from the non-overlapping window and remove the duplicates.
    \item  \emph{Seg2seq}. Seq2seq with action segmentation model has an encoder, a decoder module and attention mechanism. The encoder is a MS-TCN model with 4 stages of convolutions with each stage having 5 layers of dilated residual convolutions outputting 64 channels. The decoder is a LSTM with 256 dimensional hidden representation. We also used a single-headed attention mechanism  with a multi-layer perceptron producing 64 dimensional representation. We first train an ASRF segmentation model to obtain raw frame-wise prediction probability without refinement, with the 4 stage MS-TCN architecture (4 stacked single stage TCNs). We  treat the frame-wise softmax probabilities as the input to the seq2seq model. Seq2seq model has an encoder module and a decoder module. During training, we input the entire input sequence to model without windowing it. The seq2seq model is trained with a primitive level cross-entropy loss. During inference,we input the entire input sequence to model without windowing it and remove the duplicates.
\end{itemize}



\section{StrokeRehab dataset description}
\subsection{Cohort selection} \label{app:sub_sec_cohort_selection}
Eligibility was determined by electronic medical records, patient self-report, and physical examination. Patients were included if they were $\geq$ 18 years old, premorbidly right-handed, able to give informed consent, and had unilateral motor stroke that resulted in arm weakness with Medical Research Council score < 5/5 in any major muscle group. Patients were excluded if they had traumatic brain injury; any musculoskeletal or non-stroke neurological condition that interferes with motor function; contracture at the shoulder, elbow, or wrist;  moderate arm dysmetria or truncal ataxia; visuospatial neglect; apraxia; global inattention; or legal blindness. A trained assessor quantified arm impairment with the upper extremity Fugl-Meyer Assessment, where a maximum score of 66 signifies no impairment~\citep{fm}. Patients were moderately to mildly impaired (scores 26-65). Table~\ref{tab:demographics-table} describes the demographic and clinical characteristics of the patients.
\begingroup
\renewcommand*{\arraystretch}{1.1}
\begin{table}[h]
\centering
\begin{center}
\begin{tabular}{|c | c | c |}
\hline
 & Training set & Test set  \\
\hline
n & 33 & 8\\
\hline
Age (in years) & 56.3 (21.3-84.3) & 60.9 (42.6-84.3)\\
\hline
Gender (Female : Male) & 18 F : 15 M & 4 F : 4 M \\
\hline
Time since stroke (in years) & 6.5 (0.3-38.4) & 3.1 (0.4-5.7)\\
\hline
Paretic side (Left : Right) & 18 L : 15 R & 4 L : 4 R \\
\hline
Stroke type & &  \\
(Ischemic : Hemorrhagic) & 30 I : 3 H & 8 I : 0 H \\
\hline
Fugl-Meyer Assessment score & 48.1 (26-65)& 49.4 (27-63)\\
\hline
\end{tabular}
\end{center}
\caption{Demographic and clinical characteristics of the patients in the cohort. Mean and ranges in parentheses are shown. The cohort is divided into a training set and a test set of mildly and moderately-impaired patients. There is no overlap of patients between the training and test set. 
}
\label{tab:demographics-table}
\end{table}
\endgroup
\subsection{Kinematic data description} \label{app_subsec:kinematic_data}
Each IMU captures 3D linear accelerations and angular velocities at 100 Hz. Angular velocities are converted to sensor-centric unit quaternions, representing the rotation of each sensor on its own axes, with coordinate transformation matrices. In addition, proprietary software (Myomotion, Noraxon) generates 22 anatomical angle values using a rigid-body skeletal model scaled to patient height. See Section~\ref{sec:app_joint_angles} for a detailed description of these angles. The resulting 76-dimensional vector thus represents the kinematic features of 3D linear accelerations, 3D quaternions, and joint angles from the upper body. As an additional feature, we included the paretic (stroke-affected) side of the patient (left or right) encoded in a one-hot vector, increasing the dimension of the feature vector to 77. Each entry (except paretic side) was mean-centered and normalized separately for each task repetition in order to remove spurious offsets introduced during sensor calibration. 

\subsection{Video feature extraction} \label{app_subsec:video_feat_ext}
We extracted and released features from raw videos using the X3D model~\citep{feichtenhofer2020x3d}. The X3D model is a 3D convolutional network designed for performing the task of video classification, efficiently. The model is pretrained on the Kinetic dataset~\citep{kay2017kinetics}, which consists of coarse actions like running, climbing, sitting, etc. Since the StrokeRehab dataset consists of subtle, sub-second actions, we fine-tuned the X3D model on the training set of StrokeRehab. For fine-tuning the model, we use video sequences as input and try to identify the primitive happening in the center frame of the videos. The fine-tuned model was then used to extract the frame-wise feature vectors from the raw videos.

\subsection{Data labeling} \label{app_subsec:data_labeling}
An expert in the functional motion taxonomy (AP) ~\citep{schambra2019taxonomy} individually trained the annotators, who underwent one month of intensive training on a series of increasingly complex activities. Once annotators labeled the training videos with high accuracy (less than 2\% errors), they were given new videos to label independently. The annotators identified and labeled functional primitives in the video, which simultaneously labeled primitives in the IMU data. To ensure consistent labeling, the expert inspected one-third of all labeled videos. Interrater reliability between the annotators and expert was high across primitives (Cohen's kappa: reaches, 0.96; repositions, 0.97; transports, 0.97; stabilizations, 0.98; idles, 0.96). One minute of recording took on average 79.8 minutes to annotate. 

\subsection{Description of the Rehabilitation Activities}  
\label{sec:app_desc_act}
Tables~\ref{tab:act_descp_1} and \ref{tab:act_descp_2} describe the activities performed by the stroke patients. 

\begingroup
\renewcommand*{\arraystretch}{1}
\begin{table}[!h]
\resizebox{\textwidth}{!}{%
\centering
\begin{tabular}{|p{2cm}|p{5cm}|p{2cm}|p{5cm}|}
\hline
\multicolumn{1}{|c|}{Activity} & \multicolumn{1}{c|}{Workspace} & \multicolumn{1}{c|}{\begin{tabular}[c]{@{}c@{}}Target\\ object(s)\end{tabular}} & \multicolumn{1}{c|}{Instructions} \\ 
\hline
Washing face & Sink with a small tub (32.3 x 24.1 x 2.5 cm³) in it and two folded washcloths on either side of the countertop, 30 cm from edge closest to patient & Washcloths, faucet handle, and tub & Fill tub with water, dip washcloth on the right side into water, wring it, wiping each side of their face with wet washcloth, place it back on countertop. Use washcloth on the left side to dry face, place it back on countertop \\ \hline
Applying deodorant & Tabletop with deodorant placed at midline, 25 cm from edge closest to patient & Deodorant (solid twist-base) & Remove cap, twist base a few times, apply deodorant, replace cap, untwist the base, put deodorant on table \\ \hline
Hair combing & Tabletop with comb placed at midline, 25 cm from edge closest to patient & Comb & Pick up comb and comb both sides of head \\ \hline
Don/doffing glasses & Tabletop with glasses placed at midline, 25 cm from edge closest to patient & Pair of glasses & Wear glasses, return hands to table, remove glasses and place on table\\ \hline
Eating & Table top with a standard-size paper plate (21.6 cm diameter) placed at midline, 2 cm from edge, utensils placed 3 cm from edge, 5 cm from either side of plate, a baggie with a slice of bread placed 25 cm from edge, 23 cm left of midline, and a margarine packet placed 32 cm from edge, 17 cm right of midline & Paper plate, fork, knife, re-sealable sandwich baggie, slice of bread, single-serve margarine container & Remove bread from plastic bag and put it on plate, open margarine pack and spread it on bread, cut bread into four pieces, cut off and eat a small bite-sized piece \\ \hline
\end{tabular}}
\caption{Description of the activities performed by the stroke impaired patients in the cohort (1/2).}
\label{tab:act_descp_1}
\end{table}

\begin{table}[t]
\resizebox{\textwidth}{!}{%
\centering
\begin{tabular}{|p{2cm}|p{5cm}|p{2cm}|p{5cm}|}
\hline
\multicolumn{1}{|c|}{Activity} & \multicolumn{1}{c|}{Workspace} & \multicolumn{1}{c|}{\begin{tabular}[c]{@{}c@{}}Target\\ object(s)\end{tabular}} & \multicolumn{1}{c|}{Instructions} \\ \hline
Drinking & Tabletop with water bottle and paper cup 18 cm to the left and right of midline, 25 cm from edge closest to patient & Water bottle (12 oz),
paper cup (4 oz) & Open water bottle, pour water into cup, take a sip of water, place cup on table, and replace cap on bottle \\ \hline
Tooth brushing & Sink with toothpaste and toothbrush on either side of the countertop, 30 cm from edge closest to patient & Travel-sized toothpaste, toothbrush with built-up foam grip, faucet handle & Wet toothbrush, apply toothpaste to toothbrush, replace cap on toothpaste tube, brush teeth, rinse toothbrush and mouth, place toothbrush back on countertop \\ \hline
Moving object on a horizontal surface & Horizontal circular array (48.5 cm diameter) of 8  targets (5 cm diameter) & Toilet paper roll wrapped in self-adhesive wrap  & Move the roll between the center and each outer target, resting between each motion and at the end \\ \hline
Moving object on/off a Shelf & Shelf with two levels (33 cm and 53 cm) with 3 targets on both levels (22.5 cm, 45 cm, and 67.5 cm away from the left-most edge) & Toilet paper roll wrapped in self-adhesive wrap  & Move the roll between the center target and each target on the shelf, resting between each motion and at the end \\ \hline
\end{tabular}}
\caption{Description of the activities performed by the stroke impaired patients in the cohort (2/2).}
\label{tab:act_descp_2}
\end{table}

\endgroup


\subsection{Description of the Joint Angles}  \label{sec:app_joint_angles}
As described in Section~\ref{sec:data}, the sensor measurements are used to compute 22 anatomical angle values using a rigid-body skeletal model scaled to the patient's height. Table \ref{tab:joint_angles} describes these joint angles in  detail.
\begingroup
\renewcommand*{\arraystretch}{1}
\begin{table}[t]
\centering
\begin{tabular}{|l|l|}
\hline
Joint/segment & Anatomical angle \\ \hline
Shoulder & \begin{tabular}[c]{@{}l@{}}Shoulder flexion/extension\\ Shoulder internal/external rotation\\ Shoulder ad-/abduction\\ Shoulder total flexion$^\ddagger$ \end{tabular} \\ \hline
Elbow & Elbow flexion/extension \\ \hline
Wrist & \begin{tabular}[c]{@{}l@{}}Wrist flexion/extension\\ Forearm pronation/supination\\ Wrist radial/ulnar deviation\end{tabular} \\ \hline
Thorax & \begin{tabular}[c]{@{}l@{}}Thoracic$^*$ flexion/extension\\ Thoracic$^*$ axial rotation\\ Thoracic$^*$ lateral flexion/extension\end{tabular} \\ \hline
Lumbar & \begin{tabular}[c]{@{}l@{}}Lumbar$^\dagger$ flexion/extension\\ Lumbar$^\dagger$ axial rotation\\ Lumbar$^\dagger$ lateral flexion/extension\end{tabular} \\ \hline
\end{tabular}
\caption{List of anatomical angles. The system uses a rigid-body skeletal model to convert the IMU measurements into joint and segment angles. $\ddagger$ Shoulder total flexion is a combination of shoulder flexion/extension and shoulder ad-/abduction. $^*$Thoracic angles are computed between the cervical vertebra and the thoracic vertebra. $\dagger$Lumbar angles are computed between the thoracic vertebra and pelvis. }
\label{tab:joint_angles}
\end{table}
\endgroup

\end{document}